\documentclass{article}
\usepackage{spconf,amsmath,graphicx}
\usepackage{xcolor}
\usepackage[francais, english]{babel}

\title{Learning to detect dysarthria from raw speech}
\name{Juliette Millet$^{1,2}$, Neil Zeghidour$^{1,3}$}
\address{$^1$ CoML, ENS/CNRS/EHESS/INRIA/PSL Research University, Paris, France\\$^2$ Laboratoire de Linguistique Formelle, CNRS/Paris Diderot University/Sorbonne Paris Cit{\'{e}}, Paris, France\\ $^3$ Facebook A.I. Research, Paris, France}

\begin{document}

\maketitle
\begin{abstract}
Speech classifiers of paralinguistic traits traditionally learn from diverse hand-crafted low-level features, by selecting the relevant information for the task at hand. We explore an alternative to this selection, by learning jointly the classifier, and the feature extraction. Recent work on speech recognition has shown improved performance over speech features by learning from the waveform. We  extend this approach to paralinguistic classification and propose a neural network that can learn a filterbank, a normalization factor and a compression power from the raw speech, jointly with the rest of the architecture. We apply this model to dysarthria detection from sentence-level audio recordings. Starting from a strong attention-based baseline on which mel-filterbanks outperform standard low-level descriptors, we show that learning the filters or the normalization and compression improves over fixed features by $10\%$ absolute accuracy. We also observe a gain over OpenSmile features by learning jointly the feature extraction, the normalization, and the compression factor with the architecture. This constitutes a first attempt at learning jointly all these operations from raw audio for a speech classification task.
\end{abstract}
\section{Introduction}
\label{sec:intro}

\begin{figure}[htb]

\begin{minipage}[b]{1.0\linewidth}
  \centering
  \centerline{\includegraphics[width=7.5cm]{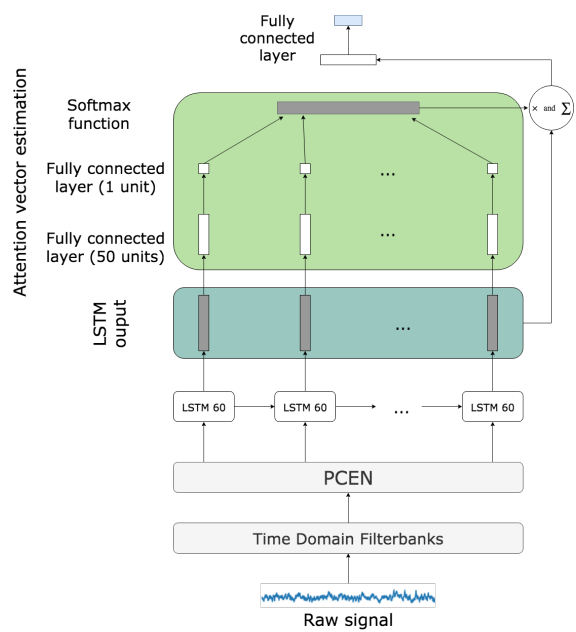}}
  
\end{minipage}
\caption{Proposed pipeline that learns jointly the feature extraction, the compression, the normalization and the classifier.}
\label{model}
\end{figure}

Learning from speech still relies on handcrafted, fixed features on which a classifier can be trained. This differs from a field like computer vision which now widely uses end-to-end models trained on raw pixels, that are typically processed by learnable convolutional operations \cite{cnn_pixels1, cnn_pixels2, cnn_pixels3}. Speech features typically contain spectral representations, such as mel-filterbanks or MFCCs, and/or low-level informations \cite{opensmile}, such as zero-crossing rate or harmonics-to-noise ratio. They are chosen to model a broad range of linguistic and paralinguistic information. Training a classifier from these fixed coefficients requires performing a feature selection step, which has the limitation that it cannot retrieve useful information that would have been lost in the feature computation. Recent research has shown improvement when replacing fixed speech features by a learnable frontend, for tasks such as speech recognition \cite{zeg2}, speaker identification \cite{cnn_speaker} or emotion recognition \cite{adieufeat}. In this work, we propose to apply such end-to-end systems to another paralinguistic task: the detection of dysarthria from speech recordings. There is a growing interest in automatically extracting information from speech for health care \cite{speech_health_1, speech_health_2, speech_health_3}, and unlike a feature-driven approach that would require testing various combinations of fixed features, we implement a system that can directly process raw speech and learn relevant features jointly with the dysarthria classifier, such that they will be optimal for the task.

The TORGO database \cite{TORGO_desc1} is a collection of annotated speech recordings and articulatory measurements from speakers with cerebral palsy (CP) or amyotrophic lateral sclerosis (ALS), as well as control patients. \cite{TORGO_adapt, TORGO_adapt2, TORGO_adapt3} have used this database to provide speech recognition systems with robustness to dysarthria. \cite{TORGO_detect} trains various linear classifiers on TORGO and the NKI CCRT corpus \cite{NKI} to detect dysarthria. More recently, \cite{TORGO_detect2} has trained fully connected neural networks to classify the severity of the disease, using TORGO and the UASPEECH \cite{UASPEECH} database. All these models are trained on standard low-level features. In this work we show that dysarthria detection benefits significantly from learning directly from the raw waveform. 

Previous work has explored learnable alternatives to speech features that rely on a similar computation to spectral representations \cite{hoshen2015speech, sainath2015learning, Ghahremani2016AcousticMF, zeg1, zeg2}. These approaches learn convolutions that are then passed through a non-linearity, eventually a pooling operator and then a log compression to replicate the dynamic range compression typically performed on spectrograms or mel-filterbanks. This compression function remains fixed and is chosen beforehand, which could impact the final performance, as various compression functions including logarithm, cubic root, or $10th$ root have been previously showed to perform better depending on the task (see Table 2 of \cite{tenth_root}). A second fixed component is the mean-variance normalization of speech features. \cite{zeg2} integrates this normalization into the neural architecture, but keeps it fixed during training. \cite{PCEN} introduces a computational block, the Per Channel Energy Normalization (PCEN) that can learn a compression and a normalization factor per channel, and can be integrated into a neural network on top of speech features. It has since then been used in production speech recognition systems \cite{baidu_pcen}.

In this work, we start from an attention-based model on mel-filterbanks, which already outperforms an equivalent model trained on low-level descriptors (LLDs). Our experiments show that by training a PCEN block on top of mel-filterbanks or replacing them by learnable time-domain filterbanks from \cite{zeg1}, we get a gain in accuracy around $10\%$ in absolute when training an identical neural network for dysarthria detection. Finally, by combining time-domain filterbanks and PCEN we propose the first audio frontend that can learn features, compression and normalization jointly with a neural network using backpropagation.

\section{MODEL}
\label{sec:model}
\subsection{Time-Domain filterbanks}
As the first step of our computational pipeline, we use Time-Domain filterbanks from \cite{zeg1}. Time-Domain filterbanks are neural network layers that take the raw waveform as input. They can be initialized to replicate mel-filterbanks, and then learnt for the task at hand. 
The standard computation of mel-filterbanks relies on passing a spectrogram through a bank of frequency domain filters. More formally, the $n^{th}$ mel-filterbank of a signal in $t$ is:
\begin{align*}
    M^nx(t) = \frac{1}{2\pi}\int |\hat{x}_t(w)|^2|\hat{\psi}_n(w)|^2 \, \mathrm dw .
\end{align*}
where $x_t(u)=\phi(t-u)x(u)$ is the waveform windowed with an Hanning function $\phi$ centered in $t$, $(\psi_n)_{n=1...N}$ the N melfilters and $\hat{f}$ denotes the Fourier transform of $f$.

\cite{mallat} shows that these coefficient can be approximated in the time domain by the following computation, referred as the first order scattering transform:
\begin{align*}
    M^nx(t) \approx |x * \varphi_n|^2*|\phi|^2(t) .
\end{align*}
where $(\varphi_n)_{n=1...N}$ are Gabor wavelets defined in \cite{zeg1} such that $|\hat{\varphi}_n|^2 \approx|\hat{\psi}_n|^2$. \cite{zeg1} shows that this computation can be implemented as neural network layers, referred as Time-Domain filterbanks (TD-filterbanks). The waveform goes through a complex-valued convolution, a modulus operator and the a convolution with a lowpass-filter (the squared hanning window) that performs the decimation. When not combined with PCEN, a log-compression is added on top of TD-filterbanks after adding $1$ to their absolute value to avoid numerical issues. Table \ref{tab:tdf} shows the detailed layers.
\begin{table}
\begin{center}
\begin{tabular}{|c|c|c|c|c|}
  \hline
Layer type   & Input  & Output   & Width & Stride \\ 

  \hline
 Conv. 1D & 1 & 128 & 400 & 1\\
 \hline
 Modulus  & 128 & 64 & - & - \\
 \hline
Square & - & - & - & -\\
  \hline
 Conv. 1D  & 64 & 64 & 400 & 160\\
  \hline
 Absolute value & - & - & - & -\\

 \hline
Add 1, log & - & - & - & -\\
   
 \hline
\end{tabular}
\caption{Description of the neural network layers used to compute time-domain filterbanks. The parameters are chosen to replicate 64 mel-filterbanks of window size 25ms and stride 10ms at 16kHz.}
\label{tab:tdf}
\end{center}
\end{table}

Following \cite{zeg1}, the first 1D convolution filters are initialized with Gabor wavelets, to replicate mel-filterbanks, and are then learnt at the same time as the rest of the model. The second convolution layer is kept fixed as a squared hanning window to perform lowpass filtering.

\subsection{Per Channel Energy Normalization}
Per Channel Energy Normalization (PCEN) is a learnable component introduced in \cite{PCEN} which computes parametrized normalization and compression. It replaces the log-compression and the mean-variance normalization. With $E(t,f)$ the value of the feature $f$ at time $t$, the computation of PCEN is:
\begin{align*}
    PCEN(t,f) = (\frac{E(t,f)}{(\epsilon + M(t,f))^\alpha} + \delta)^r - \delta^r .
\end{align*}
$M(t,f)$ is a moving average of the feature $f$ along the time axis, defined as:
\begin{align*}
    M(t,f) = (1-s)M(t-1, f) + sE(t,f) .
\end{align*}
$\alpha$ controls the strength of the normalization, the exponent $r$ (typically in $[0, 1]$) defines the slope of the compression curve, $s$ sets the spread of the moving average, and $\epsilon$ is a small scalar used to avoid division by zero. By backpropagation, we learn $\alpha$, $r$, and $\delta$ with the rest of the model to obtain a compression and a normalization that fit the task at hand.

\subsection{LSTM and Attention model}
The output of the learnable frontend is fed to an attention-based model \cite{attention}, that contains one LSTM layer of hidden size 60 followed by an attention mechanism, inspired by \cite{ART5}. The attention mechanism consists of two fully connected layers, of 50 and 1 unit respectively, and a softmax layer, that are applied to each output of the LSTM. The vector obtained is used to weight a linear combination of the LSTM outputs, that goes throught another fully connected layer of size the number of labels considered. The detailed architecture is shown in Figure \ref{model}. In \cite{ART5}, this model reaches state-of-the-art performance when trained for emotion recognition on mel-filterbanks, which motivated using it for the paralinguistic task of dysarthria detection.

\section{EXPERIMENTAL SETUP}
\label{sec:setup}
We carry experiments on the TORGO database \cite{TORGO_desc}. It consists of sound recordings, sampled at 16kHz, from speakers with either cerebral palsy or amyotrophic lateral sclerosis, which are two of the prevalent causes of speech disability or dysarthria. Similar data for a control set of subjects is also available. Along with sound recordings, TORGO contains 3D articulatory features that we did not use.

There are five groups of people: the control group not affected by the disease, and 4 other groups of affected people, classified by the severity of the disease. Each person recorded has a code name, \textbf{F} is for female, \textbf{M} is for male, while \textbf{C} is for control, followed by an identification number. A random split of the database would result in similar speakers in training, validation, and test sets, that could reduce the task to a speaker identification task. To avoid this confounding factor, we split the database to have a good repartition of the different severities among the training, validation and test set, while having no common speakers between the different sets (see Table \ref{tab:rep} for the detailed split). 

\begin{table}
\begin{center}
\begin{tabular}{|c|c|c|}
\hline
\textbf{Training}& \textbf{Validation} & \textbf{Test}\\
 \hline
FC02, F03 (VL), & MC02, FC01, & FC03, F04 (VL),  \\
F01 (L), MC04,  & M03 (VL), M01 (M) & MC01, M05 (L),\\
 MC03, M02 (M)& &  M04 (M)\\
 \hline
 3182 C, 1382 D & 950 C, 802 D & 2103 C, 997 D\\
 \hline
\end{tabular}
\caption{Speakers and number of recordings per set: C is control and D is dysarthric, the severity of each person is indicated after their ID: VL is Very Low, L is Low, and M is Medium}
\label{tab:rep}
\end{center}
\end{table}

After studying the database we decided to pad the recordings so they all last 2.5s. We extracted some typical low level descriptors (LLDs) from it to have a first baseline. We use the OpenSmile toolkit \cite{opensmile}, with the configuration of the Interspeech 2009 Emotion Challenge  \cite{interspeech2009}. For each 25ms window of the recordings (strided by 10ms), 32 features are extracted (12 MFCCs, root mean square energy, zero-crossing rate, harmonics-to-noise ratio, $F_0$ and their $\Delta$).

Our second baseline takes as input mel-filterbanks. We pre-emphase the sound signals with a factor of $0.97$. 64 mel-filterbanks are computed every 25ms with a stride of 10ms and passed through a log-compression. To evaluate our learnable frontend in a comparable setting, we design them with the same number of filters, window size and stride (see Table \ref{tab:tdf}).

For the PCEN layer, we take $\epsilon=10^{-6}$ and $s=0.5$, both fixed, and we only consider the absolute value of $r$. We initialize $r$, $\alpha$ and $\delta$ at $0.5$, $0.98$ and $2.0$ respectively. All models are trained with a stochastic gradient descent with momentum ($0.98$) and batch size 1, with a learning rate of $0.001$. 

We use the Unweighted Average Recall (UAR) to evaluate our results. The UAR of a model is the mean of its accuracy for each label. It is a better metric when dealing with unbalanced datasets than the accuracy, since it is reweighting the results depending on the size of each class. It has been widely used in unbalanced settings such as the Emotion Recognition challenge \cite{interspeech2009}. We use the validation set for hyperparameter selection and early stopping.

\section{RESULTS}
\label{sec:results}
Table \ref{tab:results} shows the UAR on the validation and test sets. All the results are the mean UAR obtained over three runs with different random initialization. We do not compare them to previously published results \cite{TORGO_detect, TORGO_detect2} as they use additional data and/or perform a different task.
\begin{table}
\begin{center}
\begin{tabular}{|c|c|c|}
\hline
\textbf{Input data} &  \textbf{UAR \% val.} & \textbf{UAR \% test}\\
 
 \hline
 LLDs & 	64.8 $\pm$ 1.2  &	65.5 $\pm$ 3.6\\
 mel-filterbanks &  79.9 $\pm$ 6.3  & 72.4 $\pm$ 3.0 \\ 
 mel-filterbanks + mvn &  63.5 $\pm$ 1.7  & 70.3 $\pm$ 2.9 \\
 mel-filterbanks + PCEN & 76.0 $\pm$ 6.1 &	79.7 $\pm$ 3.8\\
 Time-Domain filterbanks &  93.7 $\pm$ 1.2 & 82.4 $\pm$ 0.4\\
 \hline
\end{tabular}
\caption{UAR (\%) of the attention-based model trained over different features or learnable frontends. The UAR is averaged over 3 runs and standard deviations are reported.}
\label{tab:results}
\end{center}
\end{table}
The attention based-model trained on LLDs features reaches an accuracy of $66\%$ and is our baseline system. Replacing LLDs by mel-filterbanks improves the performance by $6\%$ in absolute. Adding a fixed mean-variance normalization step (mvn) brings the models to over-fitting, and thus the UAR decreases of $2\%$. However, we observe that replacing the fixed log-compression and mean-variance normalization step by a learnable PCEN layer improves the UAR of the models of $7\%$ compared to the unnormalized mel-filterbanks. Moreover, an even bigger increase is noticed when replacing mel-filterbanks by equivalent TD-filterbanks ($10 \%$ in absolute). We can emphasize the fact that using the TD-filterbanks also leads to a more stable learning process, as the standard deviation along different runs is considerably lower.

\begin{figure}[htb]
\begin{minipage}[b]{0.9\linewidth}
  \centering
  \centerline{\includegraphics[width=8.0cm]{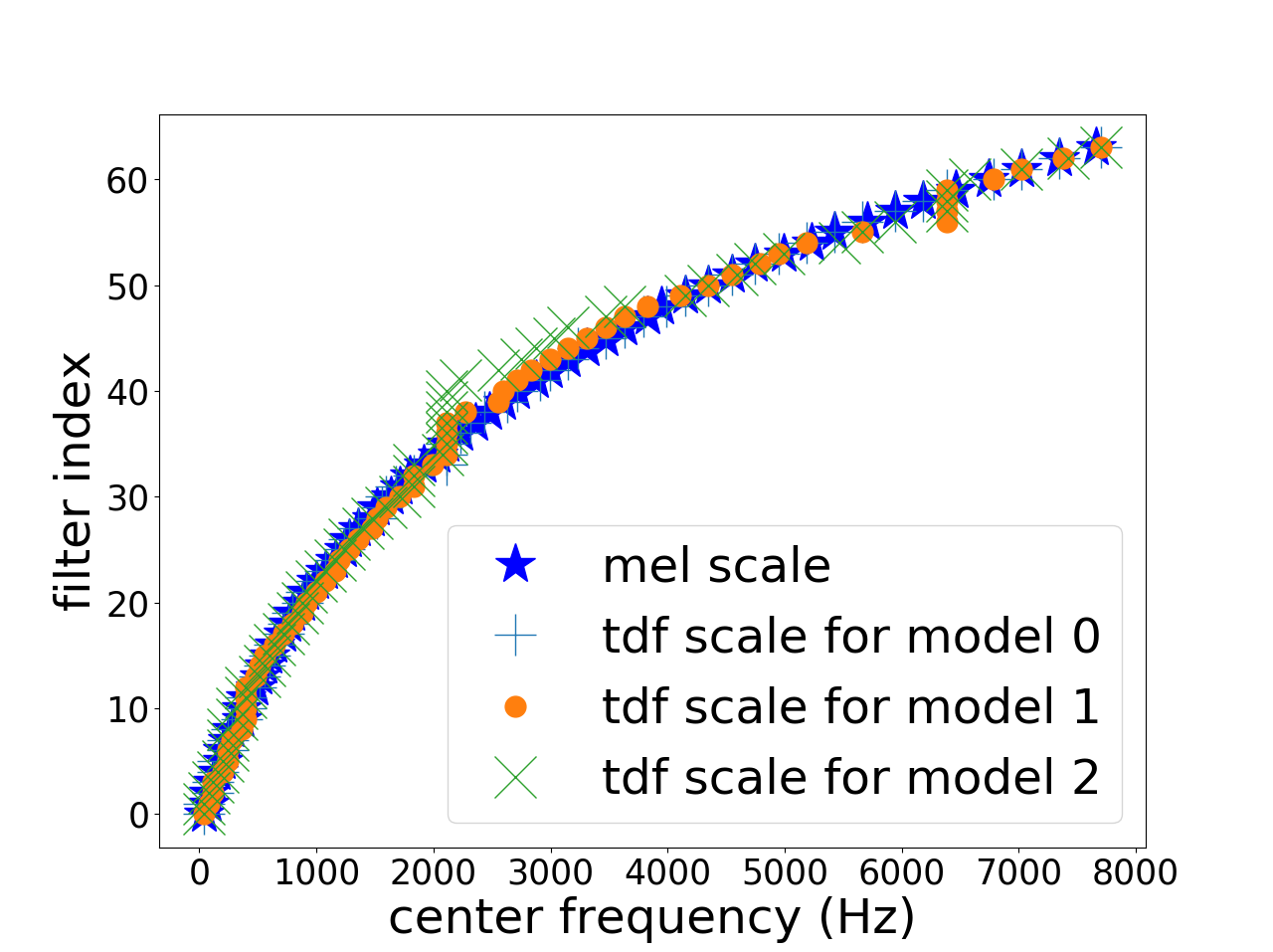}} 

\end{minipage}
\caption{New scales obtained by three independent models using TD-filterbanks, compared to mel scale. The center frequency is the frequency for which a filter is maximum.}
\label{fig:new_scale}
\end{figure}

\begin{figure}[htb]
\begin{minipage}[b]{0.9\linewidth}
  \centering
  \centerline{\includegraphics[width=7.0cm]{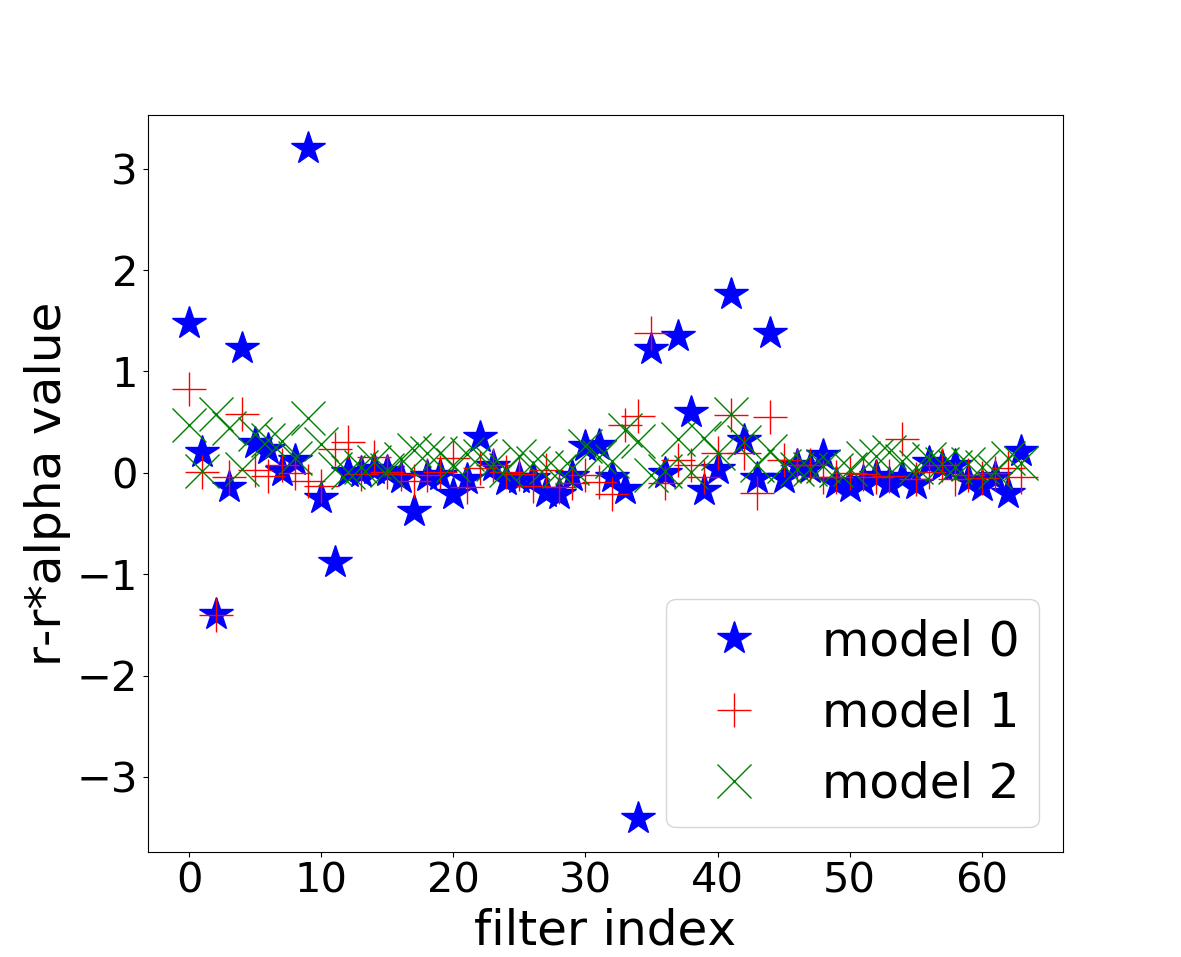}} 

\end{minipage}
\caption{Approximation of the compression exponent obtained for the PCEN layer learned on mel-filterbanks.}
\label{fig:new_compression}
\end{figure}

When studying the new scale learned by the TD-filterbanks (see Figure \ref{fig:new_scale}) we notice that the filters tend to focus around $2000Hz$ and $6500Hz$, which suggests that either those frequencies are crucial to identify dysarthria, or the model might exploit a bias in the dataset. In Figure \ref{fig:new_compression} we observe that the parameters learned by the PCEN layer reproduce similar schemes from one model to another, and that the learnt compression varies between filters, unlike a log-compression which is applied equivalently to all channels.

\subsection{Fully learnable frontend}

As we observe independent gains from either learning the features or learning the compression-normalization, we explore in our final experiments learning jointly all these operations. We remove the log-compression step of Time-Domain filterbanks and replace it by a PCEN layer. We use three settings: one for which $r$, $\alpha$ and $\delta$ are learned, the second one with only r learned, and finally the last one for which only $\alpha$ is learned. If a parameter is not learned, it is fixed to its initial value (specified in Section \ref{sec:setup}). Table \ref{tab:results_tdf_pcen} shows that learning only the normalization exponent gives worse results than the models trained on LLDs. However, we notice that the model learning $r$, $\alpha$ and $\delta$, and the one only learning $r$ match the models using mel-filterbanks.

\begin{table}
\begin{center}
\begin{tabular}{|c|c|c|}
\hline
\textbf{Input data} & \textbf{UAR \% val.} & \textbf{UAR \% test} \\
 \hline
TD-f + PCEN &  $72.3 \pm 1.5$ & $74.8 \pm 1.1$\\
\hline
TD-f + PCEN only r  & $74.6 \pm 2.9$ &	$76.4 \pm 1.8$\\
\hline
TD-f + PCEN only $\alpha$  & $66.6 \pm 1.4$ &	$63.3 \pm 8.2$ \\
 \hline
\end{tabular}
\caption{UAR (\%) of the attention-based model trained over different fully learnable frontends. The UAR is averaged over 3 runs and standard deviations are reported.}
\label{tab:results_tdf_pcen}
\end{center}
\end{table}

\section{CONCLUSION}
\label{sec:conclusion}
This paper presents a fully learnable audio frontend, combining Time-Domain filterbanks and Per Channel Energy Normalization. It is the first time that a model is developed with the ability to learn the extraction, compression and normalization of the features from the raw waveform, jointly with a classifier. We apply it to dysarthria detection, and show that replacing fixed features by learnable frontends leads to an increase in performance of the models for this task, consistently with previous results on other linguistic and paralinguistic tasks. Learning only the Time-Domain filterbanks or the PCEN parameters gives better results than learning them jointly, but learning both still gives similar to better performance than using fixed features, which constitutes a proof of concept for fully learnable audio frontends.

\begin{small}
\bibliographystyle{IEEEbib}
\bibliography{main}
\end{small}
\end{document}